\documentclass[11pt,a4paper]{article}
\usepackage{acl2015}
\usepackage{times}
\usepackage{array}
\usepackage{url}
\usepackage{latexsym}
\usepackage{multirow}
\usepackage{graphicx}
\usepackage{epstopdf}

\usepackage[labelsep=colon]{caption}

\title{Dependency Parsing with LSTMs: An Empirical Evaluation}

\author{Adhiguna Kuncoro$^{\spadesuit}$ ~ Yuichiro Sawai$^{\diamondsuit}$ ~ Kevin Duh$^{\heartsuit}$ ~ Yuji Matsumoto$^{\diamondsuit}$\\
$^{\spadesuit}$School of Computer Science, Carnegie Mellon University, Pittsburgh, PA, USA \\
$^{\diamondsuit}$Graduate School of Information Science, Nara Institute of Science and Technology, Japan \\
$^{\heartsuit}$HLTCOE, Johns Hopkins University, Baltimore, MD, USA\\
{\small \tt akuncoro@cs.cmu.edu, \{sawai.yuichiro.sn0,matsu  \}@is.naist.jp, kevinduh@cs.jhu.edu}
}

\date{}

\begin{document}
\maketitle
\begin{abstract}
We propose a transition-based dependency parser using Recurrent Neural Networks with Long Short-Term Memory (LSTM) units. This extends the feedforward neural network parser of \newcite{chenmanning14} and enables modelling of entire sequences of shift/reduce transition decisions. On the Google Web Treebank, our LSTM parser is competitive with the best feedforward parser on overall accuracy and notably achieves more than 3\% improvement for long-range dependencies, which has proved difficult for previous transition-based parsers due to error propagation and limited context information. Our findings additionally suggest that dropout regularisation on the embedding layer is crucial to improve the LSTM's generalisation.


\end{abstract}

\section{Introduction}

Two complementary approaches to transition-based dependency parsers have emerged recently. 
The {\em feature engineering} approach relies on hand-crafted feature templates to model interactions between sparse lexical features. While manually crafting these feature templates requires substantial expertise and extensive trial-and-error, this approach has led to state-of-the-art parsers in many languages \cite{BM06,ZN11}.

In contrast, the {\em neural network} approach enables automatic learning of feature combinations through  non-linear hidden layers and mitigates sparsity issues by sharing similar low-dimensional distributed  representations for related words \cite{Bet03}.

In this work, we explore new model architectures under the neural network approach. In particular, we address the issue that the feedforward architecture of the Chen and Manning parser performs training on each oracle configuration \textit{independently} of one another, disregarding the fact that the oracles for each training sentence represent a \textit{whole sequence} of intertwined decisions. Our proposed extension uses a Recurrent Neural Network (RNN) with Long Short-Term Memory (LSTM) units \cite{HS97}. At each time step of the transition system, the LSTM has theoretical access to the \textit{entire history} of past decisions (i.e. shift or reduce). LSTMs are naturally suited for modelling sequences and have shown promising results in e.g. machine translation \cite{Set14} and text-vision modelling \cite{Vet141}.

\par We particularly focus on the LSTM's performance in identifying \textit{long-range dependencies}. Such dependencies have proved difficult for most greedy transition-based parsers \cite{MN07}, including our feedforward baselines, that train on each oracle independently. This difficulty can be attributed to two main reasons: 1) most long-range dependencies are ambiguous, while the classifiers only have access to a limited context window, and 2) longer arcs are constructed after shorter arcs in transition-based parsing, thus increasing the chance of error propagation. In contrast, our LSTM has the key abilities of modelling  whole sequences of training oracles and memorise all past context information, both of which are likely beneficial for longer dependencies.

\par Despite the LSTM's theoretical advantages, in practice it is more prone to overfitting than the feedforward architecture, even with the same number of parameters. An additional contribution of this work is an empirical investigation that suggests that dropout \cite{Set142}, particularly when applied to the embedding layer, substantially improves the LSTM's generalisation ability regardless of hidden layer size. 


\section{LSTM Parsing Model}

\begin{figure*}[t]
\setlength\belowcaptionskip{-1em}
\centering
\captionsetup{justification=centering}
  \begin{center}
    \includegraphics[width=0.85\linewidth]{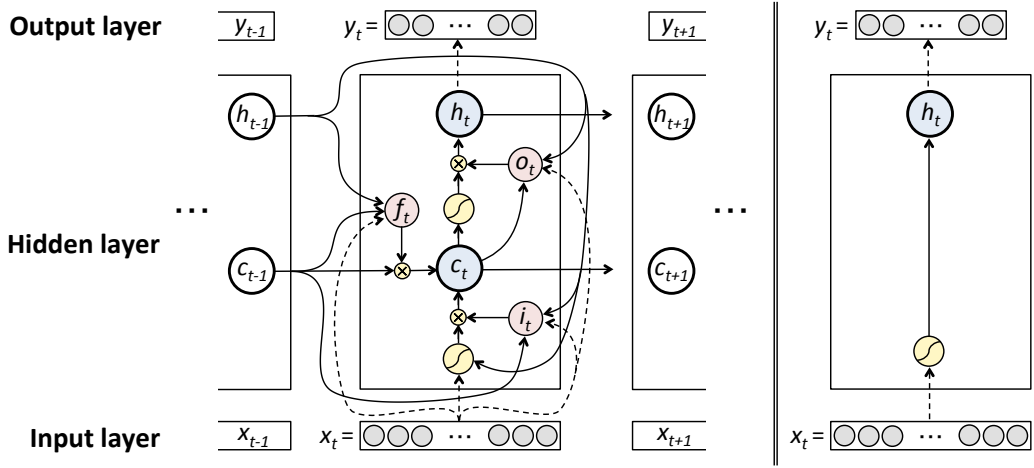}
  \end{center}
  \setlength\abovecaptionskip{-.7truemm}
  \caption{Left: Our LSTM architecture. Right: Feedforward architecture in \newcite{chenmanning14}. Connections with dropout are denoted by dashed lines.}
\label{fig:model}
\end{figure*}

\subsection{Baseline Model}
Our model is an extension to \newcite{chenmanning14},
which uses a feedforward neural network to predict the next transition of an arc-standard system.
In arc-standard, a configuration consists of a buffer $B$ (holding the input words), a stack $S$ (holding the partial parse trees), and a set of dependency arcs $A$.
The parse tree is built by successively making one of these transitions: 
\vspace{-.5em}
\begin{itemize}
\item SHIFT: move the next word on $B$ to $S$
\vspace{-.4em}
\item LEFT/RIGHT-ARC($L$): add left/right arc with label $L$ between top two words on $S$
\end{itemize}
\vspace{-.5em}

The features $x_t$ is a concatenation of embeddings from the top 3 words in $S$ and $B$, first and second left-/right-most children of the top two words on $S$, and the leftmost of leftmost / rightmost of rightmost children of the top two words on $S$. At each configuration at time $t$, the neural network first computes the hidden layer $h_t$ from the input $x_t$ (applying a non-linear function $f$), then calculates the probability of each transition in the output vector $y_t:$\footnote{The dimension of $x_t$ is 2400 in experiments, and dimension of $y_t$ is $2|L|+1$, the number of possible transitions where $|L|$ = number of dependency label types.} 
\vspace{-1.0em}
\begin{eqnarray*}
  h_{t}\!\!\! &=& \!\!\! f (W_{xh} D(x_{t})) \\[-.2em]
  y_{t}\!\!\! &=& \!\!\! \mathrm{softmax} (W_{hy} D(h_{t}))
\end{eqnarray*}
$D(\cdot)$ is a dropout operator, which randomly sets elements to 0 with probability $p_{drop}$.

\vspace{-.5em}
\subsection{Our LSTM Model}
Our LSTM model (shown in Figure~\ref{fig:model}) uses the same $x_t$ features as \newcite{chenmanning14}, but importantly adds new inputs based on past information (such as $h_{t-1}$). The addition of previous state leads to recurrence and enables modelling and training of the entire sequence of transitions. 

While recurrence may cause the ``vanishing gradient'' problem \cite{Bet94},
the LSTM architecture solves this by introducing memory cells $c_t$ that could store information over long time intervals and keep gradients from diminishing. Input gates $i_t$ control what is stored in a memory cell $c_t$, and output gates $o_t$ control whether the stored information is used in further computations. This allows information from the beginning of the sentence to influence transition actions at the end of the sentence. Forget gates $f_t$ are used to erase the information in the current memory cell. 

The following equations describe our LSTM model with peephole connections \cite{Get02}, as set forth by \newcite{G13}, and apply dropout similar to \newcite{Zet14}.
\vspace{-.6em}
\begin{eqnarray*}
  i_{t}\!\!\! &=& \!\!\! \sigma (W_{xi} D(x_{t}) + W_{ci} c_{t-1} + W_{hi} h_{t-1}) \\[-.25em]
  f_{t}\!\!\! &=& \!\!\! \sigma (W_{xf} D(x_{t}) + W_{cf} c_{t-1} + W_{hf} h_{t-1}) \\[-.25em]
  c_{t}\!\!\! &=& \!\!\! f_{t} c_{t-1}\! + \!i_{t} \ tanh (W_{xc} D(x_{t})\! + \!W_{hc} h_{t-1}\!) \\[-.25em]
  o_{t}\!\!\! &=& \!\!\! \sigma (W_{xo} D(x_{t}) + W_{co} c_{t} + W_{ho} h_{t-1}) \\[-.25em]
  h_{t}\!\!\! &=& \!\!\! o_{t} \ \tanh(c_{t}) \\[-.25em]
  y_{t}\!\!\! &=& \!\!\! \mathrm{softmax}(W_{hy} D(h_{t}))
\end{eqnarray*}

\vspace{-.6em}
Crucially, the LSTM not only uses input $x_t$ in its predictions for $y_t$, but also exploits values in the previous memory cell $c_{t-1}$ and hidden layer $h_{t-1}$ through the gates $i_t$, $f_t$, and $o_t$. The values of these gates are bounded between $[0,1]$ due to the sigmoid  $\sigma$, so multiplication with other components modulates what information is passed through. 

Given training sentences ${\{s_{i}\}}_{i=1}^{m}$ with gold parse trees,
our training data is a set of sequences of configurations $c_{it}$ and oracle transition actions $a_{it}$ at each time $t$ for each sentence $s_{i}$.
We maximise the log-likelihood of the oracle transition actions $a_{it}$ given by Equation (\ref{eq:ll}),
where $\theta$ is the set of parameters including word, POS, and label embeddings,
and $y_{t}(a)$ is the probability that the parser takes transition action $a$ at time $t$.
\vspace{-1em}
\begin{equation}
  L(\theta) = \sum_{i=1}^{m} \sum_{t} \log y_{t}(a_{it})
  - \frac{\lambda}{2} ||\theta||^2
  \label{eq:ll}
\end{equation}

\vspace{-1em}
We optimise by gradient backpropagation through time (BPTT) for each sentence $s_{i}$, feeding the parser with gold sequence of configurations $\{c_{it}\}_{t=1}^{|s_{i}|}$. 
When the parser reaches the final configuration, the gradients are backpropagated
from each prediction ${y_{it}}$ at time $t$ down to time $1$.
\section{Experiment}
\subsection{Experimental Settings} \label{exp_settings_large}
We conducted the experiments on the Google Web Treebank \cite{PM12}, consisting of the WSJ portion of the OntoNotes corpus and five additional web domains, with 48 dependency types. The models were trained only on the training set of the WSJ corpus, while the parameters were optimised using the WSJ dev set (i.e. no tuning using any of the web domains' dev set).


As baselines, we re-implemented the Chen and Manning parser with the same setting, including results from both the feedforward model with Tanh activation function (same activation as the LSTM) and its better-performing Cubic counterpart. 
Training was done for a maximum of 400 epochs, stopped early if no better dev UAS was found after 30 consecutive epochs.\footnote{The LSTM was trained with the Adadelta optimiser \cite{Z12}, using a decay rate of 0.95 and $\epsilon=10^{-6}$. The embeddings were similarly initialised as the feedforward baselines, while the weight connections were initialised using the same mechanism as \newcite{GB10}.
We used automatic POS tags from the Stanford bi-directional tagger \cite{Tet03}, with tagging accuracies of 97\% for the WSJ and 87-92\% for the web domains.}


\subsection{Main Result and Analysis}
\begin{figure*}[t]
\centering
\begin{minipage}{\columnwidth}
  \centering
  \includegraphics[width=.9\columnwidth]{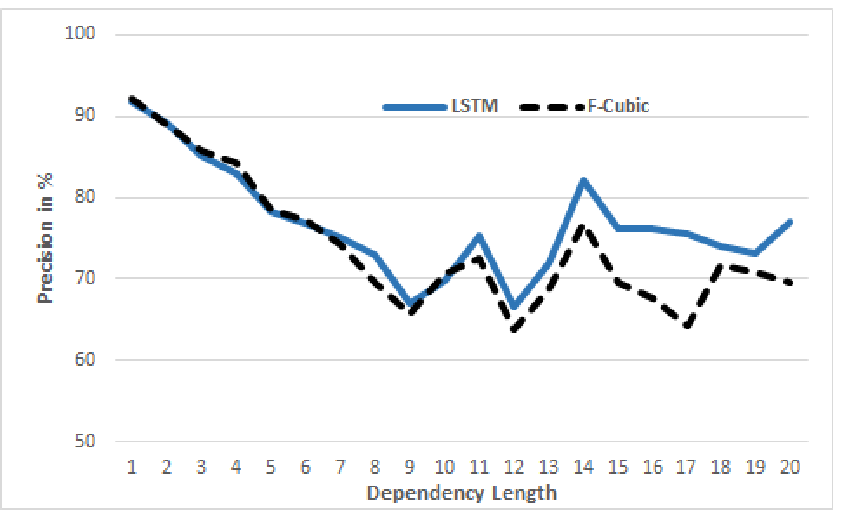}
  \caption{Precision by Dependency Length}
  \label{fig:precision}
\end{minipage}
\begin{minipage}{.95\columnwidth}
  \centering
  \includegraphics[width=.9\columnwidth]{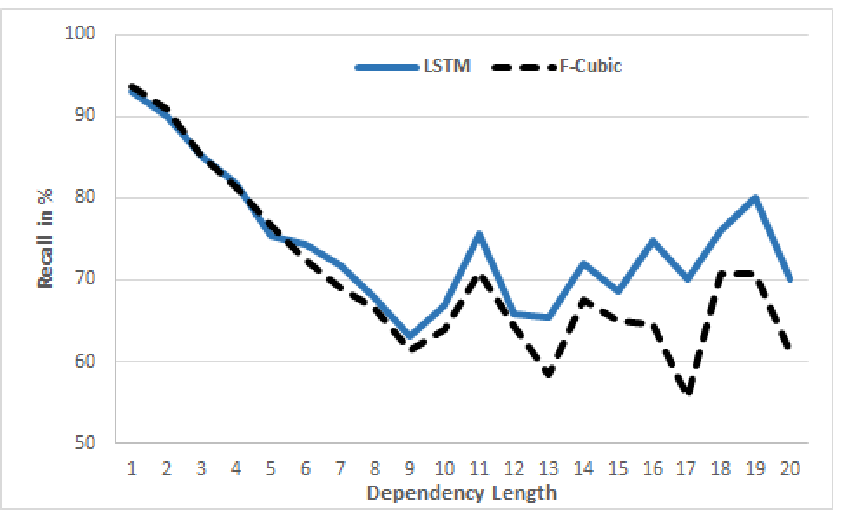}
  \caption{Recall by Dependency Length}
  \label{fig:recall}
\end{minipage}
\end{figure*}
The LAS result on the Google Web Treebank is summarised on Table \ref{full_result_table}, where F-T and F-C represent the feedforward baselines with Tanh and Cubic activations, respectively.
Our LSTM model outperforms the feedforward baseline with the same Tanh activation function (87.5 vs 86.4 on WSJ Test), while achieving competitive accuracy with the Cubic baseline. 

We furthermore investigate the models' performance on long-range dependencies, reporting the result in terms of labelled precision and recall breakdown by dependency lengths on the WSJ test set in Table \ref{tab_long_range}. This result is also plotted in Figures \ref{fig:precision} and \ref{fig:recall}. 
Despite the models' similar overall accuracy, our LSTM model outperforms the Cubic baseline by more than 3\% in both precision and recall for dependency lengths greater than 7, and that the LSTM's performance degrades more slowly as dependency length increases.


\begin{table}[t]
\centering
\begin{tabular}{|c|c|c|c|}
\hline
{\bf WSJ} & {\bf F - T} & {\bf F - C} & {\bf LSTM} \\ \hline
Dev             & 86.0        & 87.2        & {\bf 87.8} \\ \hline
Test            & 86.4        & {\bf 87.5}  & {\bf 87.5} \\ \hline \hline
{\bf Web Test}       & {\bf F - T} & {\bf F - C} & {\bf LSTM} \\ \hline
Answers         & 74.1        & {\bf 74.9}  & 74.6       \\ \hline
Emails          & 74.6        & {\bf 75.6}  & 74.4       \\ \hline
Newsgroups      & 79.3        & 79.9        & {\bf 80.2} \\ \hline
Reviews         & 76.5        & {\bf 77.2}  & 77.0       \\ \hline
Weblogs         & 80.7        & 81.1        & {\bf 81.2} \\ \hline
\end{tabular}
\caption{Google Web Treebank LAS Result}
\label{full_result_table}
\end{table}


\begin{table}[t]
\centering
\begin{tabular}{|c|c|c|c|c|}
\hline
{\bf \begin{tabular}[c]{@{}c@{}}Dep.\\ Length\end{tabular}} & \multicolumn{1}{l|}{} & F - T & F - C      & LSTM       \\ \hline
\multirow{2}{*}{1}                                          & Precision             & 91.4  & {\bf 92.2} & 91.7       \\ \cline{2-5} 
                                                            & Recall                & 93.1  & {\bf 93.6} & 93.0       \\ \hline
\multirow{2}{*}{2}                                          & Precision             & 87.9  & 88.9       & {\bf 89.3} \\ \cline{2-5} 
                                                            & Recall                & 90.3  & {\bf 91.0} & 90.1       \\ \hline
\multirow{2}{*}{3-6}                                        & Precision             & 81.7  & {\bf 83.4} & 82.6       \\ \cline{2-5} 
                                                            & Recall                & 79.2  & 81.3       & {\bf 81.4} \\ \hline
\multirow{2}{*}{7-49}                                       & Precision             & 68.1  & 70.3       & {\bf 73.5} \\ \cline{2-5} 
                                                            & Recall                & 62.6  & 65.6       & {\bf 69.5} \\ \hline
\end{tabular}
\caption{Long-range Arcs Precision \& Recall}
\label{tab_long_range}
\end{table}

\subsection{Regularisation Experiments}\label{hyperparameters}

We discover that regularisation is important for the LSTM parser, more so than feedforward architectures. Table \ref{table:with_without_dropout} compares the relative improvement due to dropout for feedforward vs. LSTM by constraining both models to have the same number of 500,000 parameters, corresponding to 50 hidden units for LSTM.
Observe that LSTM becomes competitive only with dropout. 


\begin{table}[h]
\centering
\begin{tabular}{|c|c|c|c|}
\hline
{\bf } & no dropout & with dropout & {\bf $\Delta$} \\ \hline
F-Cubic    & 89.1        & 89.5        & {\bf 0.4} \\ \hline
LSTM       & 87.4        & 89.5  & {\bf 2.1} \\ \hline
\end{tabular}
\caption{\label{table:with_without_dropout} Effect of Dropout on UAS Accuracy}
\end{table}




To investigate what kind of dropout is beneficial,  we conducted further experiments on a subset of the training data (the first 80,000 tokens of the WSJ training set).\footnote{We used the same experimental settings as in Subsection \ref{exp_settings_large} and evaluate UAS on the full WSJ dev and test set, with hidden layer size fixed at 60.}
The results of dropout and L-2 regularisation are in Table \ref{tab_regul}, along with the epoch where the best dev UAS is found. E-H and H-O indicate dropout between the embedding-hidden and hidden-output connections, respectively. 

\begin{table}[t]
\centering
\begin{tabular}{|c|c|c|c|c|}
\hline
\multicolumn{2}{|c|}{{\bf Reg　Settings}} & {\bf Dev } & {\bf Test } & {\bf Epoch} \\ \hline \hline
\multicolumn{2}{|c|}{{\bf L2 $\lambda$}}    &               &                &             \\ \hline
\multicolumn{2}{|c|}{0}                  & 80.2         & 80.0           & 42          \\ \hline
\multicolumn{2}{|c|}{$10^{-8}$}                  & 80.7          & 80.8          & 25          \\ \hline
\multicolumn{2}{|c|}{$10^{-7}$}                  & 79.9          & 80.3           & 43          \\ \hline
\multicolumn{2}{|c|}{$10^{-6}$}                  & 79.8          & 80.1           & 43         \\ \hline
\multicolumn{2}{|c|}{$10^{-5}$}                  & 80.5          & 80.4          & 46          \\ \hline
\multicolumn{2}{|c|}{$10^{-4}$}                  & {\underline{83.4}}          & {\underline{82.9}}           & 206          \\ \hline
\multicolumn{2}{|c|}{$10^{-3}$}                  & 81.6          & 81.6           & 159          \\ \hline \hline
\multicolumn{2}{|c|}{\bf Dropout $p_{drop}$}  &               &                &             \\ \hline
\multirow{3}{*}{E-H}      & 0.2          & 84.4          & 84.3          & 97          \\ \cline{2-5} 
                          & 0.4          & 85.8          & {\underline{85.7}}           & 257          \\ \cline{2-5} 
                          & 0.6          & {\underline{\bf 86.2}}         & 85.5           & 273          \\ \hline
\multirow{3}{*}{H-O}      & 0.2          & 81.8          & 81.6          & 52          \\ \cline{2-5} 
                          & 0.4          & {\underline{82.3}}          & {\underline{82.1}}          & 93          \\ \cline{2-5} 
                          & 0.6          & 81.9          & 81.7          & 69          \\ \hline
\multirow{3}{*}{Both}  & 0.2          & 85.4          & 85.0          & 122          \\ \cline{2-5} 
                          & 0.4          & {\underline{86.1}}         & {\underline{\bf 85.9}}           & 315          \\ \cline{2-5} 
                          & 0.6          & 85.3          & 85.3           & 500          \\ \hline
\end{tabular}
\caption{UAS Accuracy of Various Regularisation}
\label{tab_regul}
\end{table}

\begin{figure}[t]
\centering
  \centering
  \includegraphics[width=.9\columnwidth]{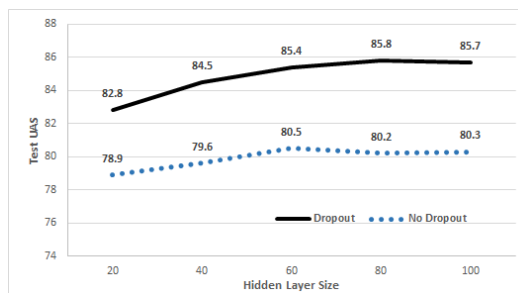}
  \caption{UAS Accuracy vs Hidden Layer Size}
\label{fig:dropout_hidden}
\end{figure}

While dropout generally results in slower convergence, the technique outperforms L-2 and significantly improves the model's accuracy by more than 6\%. Most importantly, we found input dropout to be more crucial than hidden-output dropout and achieves the same accuracy as dropout on both input and hidden layers, suggesting that our model can achieve good accuracy with input dropout alone. 
We found dropout rates between 0.4 and 0.6 to be effective.
Further, we found that dropout generally improves LSTMs regardless of model size. Figure \ref{fig:dropout_hidden} shows how dropout of 0.5 on E-H and E-O layers improve results for various hidden layer sizes. 

\section{Related work} Recently, various neural network models have achieved state of the art results in many parsing tasks and languages, including the Google Web Treebank dataset used in this paper. \newcite{Vet142} used LSTMs for sequence-to-sequence constituency parsing that makes no prior assumption of the parsing problem. For dependency parsing, \newcite{S13} presented an RNN compositional model, similar to the RNN constituency parser of \newcite{Set13}. 

More recently, the works of \newcite{Det15} and \newcite{KiperwasserGoldberg16} proposed transition-based LSTM models to automatically extract real-valued feature vectors from the parser configuration. The transition-based parser of \newcite{Det15} used a ``stack LSTM'' architecture and composition functions to obtain a continuous, low-dimensional representation of the stack to represent partial trees, along with the buffer and history of actions. Both our work and the stack LSTM model similarly used greedy decoding, although one primary difference is that we used the LSTM to form temporal recurrence over the \emph{hidden states}\footnote{We define the hidden states as the penultimate layer right before the softmax.}. We used the same feature extraction template as \newcite{chenmanning14} and replaced the feedforward connections with LSTM network, while \newcite{Det15} instead used the stack LSTM as a means to extract dense features from the parser configuration without explicit temporal recurrence.

Neural network models have also been used for structured training in transition-based parsing, achieving state of the art results on various dataset. \newcite{Wet15} used a structured perceptron model on top of a feedforward transition-based dependency parser. When augmented with tri-training method on unlabelled data, their model achieved an impressive 87\% LAS on the Web domain data of the Google Web Treebank similarly used in this work. \newcite{Zet15} used beam search and contrastive learning to maximise the probability of the entire gold \emph{sequence} with respect to all other sequences in the beam. \newcite{Aet16} similarly proposed a globally normalised model using beam search and Conditional Random Fields (CRF) loss \cite{Let01} that achieved state of the art results on the benchmark English PTB dataset. 

Our RNN parsing model is most similar with \newcite{Xet16} that used temporal recurrence over the hidden states for CCG parsing, although we use LSTMs instead of Elman RNNs. Our work additionally investigates the effect of dropout on model performance, and demonstrate the efficacy of temporal recurrence to better capture long-range dependencies.

\section{Conclusions}

We present a transition-based dependency parser using recurrent LSTM units. The motivation is to exploit the entire history of shift/reduce transitions when making predictions. This LSTM parser is competitive with the feedforward neural network parser of \newcite{chenmanning14} on overall LAS, and notably improves the accuracy of long-range dependencies. We also show the importance of dropout, particularly on the embedding layer, in improving the model's accuracy.

\section*{Acknowledgments}

We thank Graham Neubig and Hiroyuki Shindo for the useful feedback and comments.

\nocite{*}
\newpage
\bibliographystyle{acl}
\bibliography{acl2015}

\begin{thebibliography}{}

\bibitem[\protect\citename{Andor \bgroup et al.\egroup }2016]{Aet16}
Daniel Andor, Chris Alberti, David Weiss, Aliaksei Severyn, Alessandro Presta,
  Kuzman Ganchev, Slav Petrov, and Michael Collins.
\newblock 2016.
\newblock Globally normalized transition-based neural networks.
\newblock {\em CoRR}, abs/1603.06042.

\bibitem[\protect\citename{Bansal \bgroup et al.\egroup }2014]{Bet14}
Mohit Bansal, Kevin Gimpel, and Karen Livescu.
\newblock 2014.
\newblock Tailoring continuous word representations for dependency parsing.
\newblock In {\em Proceedings of ACL}.

\bibitem[\protect\citename{Bastien \bgroup et al.\egroup }2012]{Bet12}
Fr{\'{e}}d{\'{e}}ric Bastien, Pascal Lamblin, Razvan Pascanu, James Bergstra,
  Ian~J. Goodfellow, Arnaud Bergeron, Nicolas Bouchard, and Yoshua Bengio.
\newblock 2012.
\newblock Theano: new features and speed improvements.
\newblock Deep Learning and Unsupervised Feature Learning NIPS 2012 Workshop.

\bibitem[\protect\citename{Bengio \bgroup et al.\egroup }1994]{Bet94}
Yoshua Bengio, Patrice Simard, and Paolo Frasconi.
\newblock 1994.
\newblock Learning long-term dependencies with gradient descent is difficult.
\newblock {\em Trans. Neur. Netw.}, 5(2):157--166, March.

\bibitem[\protect\citename{Bengio \bgroup et al.\egroup }2003]{Bet03}
Yoshua Bengio, Réjean Ducharme, Pascal Vincent, and Christian Jauvin.
\newblock 2003.
\newblock A neural probabilistic language model.
\newblock {\em JOURNAL OF MACHINE LEARNING RESEARCH}, 3:1137--1155.

\bibitem[\protect\citename{Bergstra \bgroup et al.\egroup }2010]{Bet10}
James Bergstra, Olivier Breuleux, Fr{\'{e}}d{\'{e}}ric Bastien, Pascal Lamblin,
  Razvan Pascanu, Guillaume Desjardins, Joseph Turian, David Warde-Farley, and
  Yoshua Bengio.
\newblock 2010.
\newblock Theano: a {CPU} and {GPU} math expression compiler.
\newblock In {\em Proceedings of the Python for Scientific Computing Conference
  ({SciPy})}, June.
\newblock Oral Presentation.

\bibitem[\protect\citename{Buchholz and Marsi}2006]{BM06}
Sabine Buchholz and Erwin Marsi.
\newblock 2006.
\newblock Conll-x shared task on multilingual dependency parsing.
\newblock In {\em Proc. of CoNLL}, pages 149--164.

\bibitem[\protect\citename{Chen and Manning}2014]{chenmanning14}
Danqi Chen and Christopher~D. Manning.
\newblock 2014.
\newblock A fast and accurate dependency parser using neural networks.
\newblock In {\em Empirical Methods in Natural Language Processing (EMNLP)}.

\bibitem[\protect\citename{Collobert \bgroup et al.\egroup }2011]{Cet11}
Ronan Collobert, Jason Weston, L{\'{e}}on Bottou, Michael Karlen, Koray
  Kavukcuoglu, and Pavel~P. Kuksa.
\newblock 2011.
\newblock Natural language processing (almost) from scratch.
\newblock {\em CoRR}, abs/1103.0398.

\bibitem[\protect\citename{Duchi \bgroup et al.\egroup }2011]{Det11}
John Duchi, Elad Hazan, and Yoram Singer.
\newblock 2011.
\newblock Adaptive subgradient methods for online learning and stochastic
  optimization.
\newblock {\em J. Mach. Learn. Res.}, 12:2121--2159, July.

\bibitem[\protect\citename{Dyer \bgroup et al.\egroup }2015]{Det15}
Chris Dyer, Miguel Ballesteros, Wang Ling, Austin Matthews, and Noah~A. Smith.
\newblock 2015.
\newblock Transition-based dependency parsing with stack long short-term
  memory.
\newblock In {\em Proceedings of the 53rd Annual Meeting of the Association for
  Computational Linguistics and the 7th International Joint Conference on
  Natural Language Processing (Volume 1: Long Papers)}, pages 334--343,
  Beijing, China, July. Association for Computational Linguistics.

\bibitem[\protect\citename{Fonseca \bgroup et al.\egroup }2015]{Fet15}
Erick~R Fonseca, Avenida~Trabalhador S{\~a}o-carlense, and Sandra~M
  Alu{\'\i}sio.
\newblock 2015.
\newblock A deep architecture for non-projective dependency parsing.
\newblock {\em Proceedings of NAACL-HLT}, pages 56--61.

\bibitem[\protect\citename{Gers \bgroup et al.\egroup }2002]{Get02}
Felix~A. Gers, Nicol~N. Schraudolph, and J\"{u}rgen Schmidhuber.
\newblock 2002.
\newblock Learning precise timing with lstm recurrent networks.
\newblock {\em jmlr}, 3:115--143.

\bibitem[\protect\citename{Glorot and Bengio}2010]{GB10}
Xavier Glorot and Yoshua Bengio.
\newblock 2010.
\newblock Understanding the difficulty of training deep feedforward neural
  networks.
\newblock In {\em Proceedings of the International Conference on Artificial
  Intelligence and Statistics (AISTATS’10). Society for Artificial
  Intelligence and Statistics}.

\bibitem[\protect\citename{Graves}2013]{G13}
Alex Graves.
\newblock 2013.
\newblock Generating sequences with recurrent neural networks.
\newblock {\em CoRR}, abs/1308.0850.

\bibitem[\protect\citename{Greff \bgroup et al.\egroup }2015]{Get15}
Klaus Greff, Rupesh~Kumar Srivastava, Jan Koutn{\'{\i}}k, Bas~R. Steunebrink,
  and J{\"{u}}rgen Schmidhuber.
\newblock 2015.
\newblock {LSTM:} {A} search space odyssey.
\newblock {\em CoRR}, abs/1503.04069.

\bibitem[\protect\citename{Hochreiter and Schmidhuber}1997]{HS97}
Sepp Hochreiter and J\"{u}rgen Schmidhuber.
\newblock 1997.
\newblock Long short-term memory.
\newblock {\em Neural Comput.}, 9(8):1735--1780, November.

\bibitem[\protect\citename{Kiperwasser and
  Goldberg}2016]{KiperwasserGoldberg16}
Eliyahu Kiperwasser and Yoav Goldberg.
\newblock 2016.
\newblock Simple and accurate dependency parsing using bidirectional {LSTM}
  feature representations.
\newblock {\em CoRR}, abs/1603.04351.

\bibitem[\protect\citename{Lafferty}2001]{Let01}
John Lafferty.
\newblock 2001.
\newblock Conditional random fields: Probabilistic models for segmenting and
  labeling sequence data.
\newblock pages 282--289. Morgan Kaufmann.

\bibitem[\protect\citename{McDonald and Nivre}2007]{MN07}
Ryan McDonald and Joakim Nivre.
\newblock 2007.
\newblock Characterizing the errors of data-driven dependency parsing models.
\newblock In {\em Proceedings of the Conference on Empirical Methods in Natural
  Language Processing and Natural Language Learning}.

\bibitem[\protect\citename{Petrov and McDonald}2012]{PM12}
Slav Petrov and Ryan McDonald.
\newblock 2012.
\newblock Overview of the 2012 shared task on parsing the web.
\newblock Notes of the First Workshop on Syntactic Analysis of Non-Canonical
  Language {(SANCL)}.

\bibitem[\protect\citename{Socher \bgroup et al.\egroup }2013]{Set13}
Richard Socher, John Bauer, Christopher~D. Manning, and Andrew~Y. Ng.
\newblock 2013.
\newblock Parsing with compositional vector grammars.
\newblock In {\em Proceedings of the ACL conference}.

\bibitem[\protect\citename{Srivastava \bgroup et al.\egroup }2014]{Set142}
Nitish Srivastava, Geoffrey Hinton, Alex Krizhevsky, Ilya Sutskever, and Ruslan
  Salakhutdinov.
\newblock 2014.
\newblock Dropout: A simple way to prevent neural networks from overfitting.
\newblock {\em Journal of Machine Learning Research}, 15:1929--1958.

\bibitem[\protect\citename{Stenetorp}2013]{S13}
Pontus Stenetorp.
\newblock 2013.
\newblock Transition-based dependency parsing using recursive neural networks.
\newblock In {\em Deep Learning Workshop at the 2013 Conference on Neural
  Information Processing Systems (NIPS)}, Lake Tahoe, Nevada, USA, December.

\bibitem[\protect\citename{Sutskever \bgroup et al.\egroup }2014]{Set14}
Ilya Sutskever, Oriol Vinyals, and Quoc~V. Le.
\newblock 2014.
\newblock Sequence to sequence learning with neural networks.
\newblock {\em CoRR}, abs/1409.3215.

\bibitem[\protect\citename{Toutanova \bgroup et al.\egroup }2003]{Tet03}
Kristina Toutanova, Dan Klein, Christopher~D. Manning, and Yoram Singer.
\newblock 2003.
\newblock Feature-rich part-of-speech tagging with a cyclic dependency network.
\newblock In {\em Proceedings of the 2003 Conference of the North American
  Chapter of the Association for Computational Linguistics on Human Language
  Technology - Volume 1}, NAACL '03, pages 173--180, Stroudsburg, PA, USA.
  Association for Computational Linguistics.

\bibitem[\protect\citename{Venugopalan \bgroup et al.\egroup }2014]{Vet141}
Subhashini Venugopalan, Huijuan Xu, Jeff Donahue, Marcus Rohrbach, Raymond~J.
  Mooney, and Kate Saenko.
\newblock 2014.
\newblock Translating videos to natural language using deep recurrent neural
  networks.
\newblock {\em CoRR}, abs/1412.4729.

\bibitem[\protect\citename{Vinyals \bgroup et al.\egroup }2014]{Vet142}
Oriol Vinyals, Lukasz Kaiser, Terry Koo, Slav Petrov, Ilya Sutskever, and
  Geoffrey~E. Hinton.
\newblock 2014.
\newblock Grammar as a foreign language.
\newblock {\em CoRR}, abs/1412.7449.

\bibitem[\protect\citename{Weiss \bgroup et al.\egroup }2015]{Wet15}
David Weiss, Chris Alberti, Michael Collins, and Slav Petrov.
\newblock 2015.
\newblock Structured training for neural network transition-based parsing.
\newblock In {\em Proceedings of the 53rd Annual Meeting of the Association for
  Computational Linguistics and the 7th International Joint Conference on
  Natural Language Processing (Volume 1: Long Papers)}, pages 323--333,
  Beijing, China, July. Association for Computational Linguistics.

\bibitem[\protect\citename{Xu \bgroup et al.\egroup }2016]{Xet16}
Wenduan Xu, Michael Auli, and Stephen Clark.
\newblock 2016.
\newblock Expected f-measure training for shift-reduce parsing with recurrent
  neural networks.
\newblock In {\em Proceedings of the 2016 Conference of the North American
  Chapter of the Association for Computational Linguistics: Human Language
  Technologies}, pages 210--220, San Diego, California, June. Association for
  Computational Linguistics.

\bibitem[\protect\citename{Yamada and Matsumoto}2003]{YM03}
Hiroyasu Yamada and Yuji Matsumoto.
\newblock 2003.
\newblock Statistical dependency analysis with support vector machines.
\newblock In {\em Proceedings of the International Workshop on Parsing
  Technologies (IWPT)}.

\bibitem[\protect\citename{Zaremba \bgroup et al.\egroup }2014]{Zet14}
Wojciech Zaremba, Ilya Sutskever, and Oriol Vinyals.
\newblock 2014.
\newblock Recurrent neural network regularization.
\newblock {\em CoRR}, abs/1409.2329.

\bibitem[\protect\citename{Zeiler}2012]{Z12}
Matthew~D. Zeiler.
\newblock 2012.
\newblock {ADADELTA:} an adaptive learning rate method.
\newblock {\em CoRR}, abs/1212.5701.

\bibitem[\protect\citename{Zhang and Nivre}2011]{ZN11}
Yue Zhang and Joakim Nivre.
\newblock 2011.
\newblock Transition-based dependency parsing with rich non-local features.
\newblock In {\em Proceedings of the 49th Annual Meeting of the Association for
  Computational Linguistics: Human Language Technologies: Short Papers - Volume
  2}, HLT '11, pages 188--193, Stroudsburg, PA, USA. Association for
  Computational Linguistics.

\bibitem[\protect\citename{Zhou \bgroup et al.\egroup }2015]{Zet15}
Hao Zhou, Yue Zhang, Shujian Huang, and Jiajun Chen.
\newblock 2015.
\newblock A neural probabilistic structured-prediction model for
  transition-based dependency parsing.
\newblock In {\em Proceedings of the 53rd Annual Meeting of the Association for
  Computational Linguistics and the 7th International Joint Conference on
  Natural Language Processing (Volume 1: Long Papers)}, pages 1213--1222,
  Beijing, China, July. Association for Computational Linguistics.

\end{thebibliography}

\end{document}